\documentclass[conference]{IEEEtran}
\IEEEoverridecommandlockouts
\usepackage{cite}
\usepackage{amsmath,amssymb,amsfonts}
\usepackage{algorithmic}
\usepackage{graphicx}
\usepackage{textcomp}
\usepackage{xcolor}
\usepackage{subcaption}
\usepackage[hyphens]{url}
\def\BibTeX{{\rm B\kern-.05em{\sc i\kern-.025em b}\kern-.08em
    T\kern-.1667em\lower.7ex\hbox{E}\kern-.125emX}}
\begin{document}

\title{Multi-Lingual DALL-E Storytime}
  


\author{
\IEEEauthorblockN{Noga Mudrik}
\IEEEauthorblockA{\textit{Biomedical Engineering and}
\\ \textit{The Center of Imaging Science}
\\The Johns Hopkins University 
 \\
nmudrik1@jhu.edu}
\and

\IEEEauthorblockN{Adam S. Charles}
\IEEEauthorblockA{\textit{Biomedical Engineering, Kavli NDI,} 
\\ \textit{and The Center of Imaging Science}
\\ The Johns Hopkins University 
 \\
adamsc@jhu.edu}
}

\vspace{-8 pt}
\maketitle
\vspace{-8 pt}
\begin{abstract}
While recent advancements in artificial intelligence (AI) language models demonstrate cutting-edge performance when working with English texts, equivalent models do not exist in other languages or do not reach the same performance level. This undesired effect of AI advancements increases the gap between access to new technology from different populations across the world. This unsought bias mainly discriminates against individuals whose English skills are less developed,  i.e., children, especially from low-income families in non-English countries, and might deepen educational gaps between children from different backgrounds. 
Following significant advancements in AI research in recent years, OpenAI has recently presented DALL-E: a powerful tool for creating images based on English text prompts. While DALL-E is a promising tool for many applications, its decreased performance when given input in a different language, limits its audience and deepens the gap between populations. An additional limitation of the current DALL-E model is that it only allows for the creation of a few images in response to a given input prompt, rather than a series of consecutive coherent frames that tell a story or describe a process that changes over time.
Here, we present an easy-to-use automatic DALL-E storytelling framework that leverages the existing DALL-E model to enable fast and coherent visualizations of non-English songs and stories, pushing the limit of the one-step-at-a-time option DALL-E currently offers. 
 We show that our framework is able to effectively visualize stories from non-English texts and portray the changes in the plot over time. It is also able to create a narrative and maintain interpretable changes in the description across frames. Additionally, our framework offers users the ability to specify constraints on the story elements, such as a specific location or context, and to maintain a consistent style throughout the visualization. We believe this tool is effective for helping people, particularly children, quickly comprehend complex narrative texts, including fast-paced songs and biblical stories.
 


\end{abstract}


\section{Introduction}
\begin{figure}[h]
\begin{center}
\includegraphics[scale=0.47]{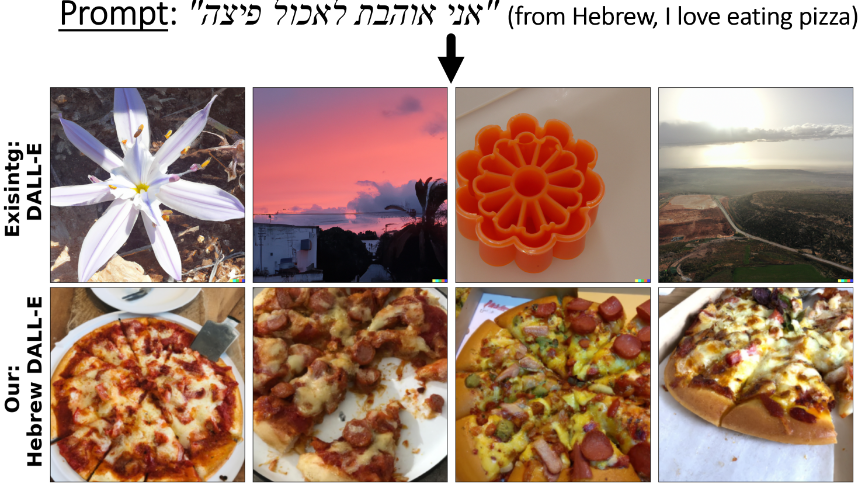}
\end{center}
   \caption{\textit{Comparing the current DALL-E model and our augmented multi-lingual DALL-E for a Hebrew prompt (translation: "I love eating pizza"). While the current available DALL-E did not throw an error when fed with a Hebrew prompt (top row), it is apparent it was not able to capture its content, in contrast to our results (bottom row).}}
\label{fig:pizza}
\vspace{-11 pt}
\end{figure}

Recent years observed a dramatic improvement in various AI capabilities in different domains, including state-of-the-art retrogressive language models ~\cite{brown2020language,radford2019language}, or models for generating images from text \cite{saharia2022photorealistic, ramesh2021zero, mansimov2015generating, zhu2007text}. 
While these tools can be an asset for promoting technology, science, and education, a natural bias arises in these systems from their development on English texts by native-English speakers, making their use focused on English-based applications only. This bias makes their use more challenging for non-English speakers, especially among groups who have minimal knowledge of English, in particular young non-English speakers children. 
Providing state-of-the-art models in every language is not feasible due to the large amount of computational resources needed for model training that would be required. However, using advanced English language models to process non-English text is a promising avenue to increase accessibility to AI tools for a variety of populations and is an important step toward ensuring more equitable access to cutting-edge tools.
OpenAI's DALL-E model \cite{ramesh2021zero} has recently been drawing increasing attention as one of the promising text-to-image models.
While the existing model does not throw an error when fed with a non-English prompt, it is not able to visualize meaningful images of the given text (e.g., Fig.~\ref{fig:pizza}).

An additional way to enhance AI models for children's education is by enabling them to tell stories or describe processes that change over time.
While the existing DALL-E model is effective for visualizing single-frame scenarios, it is currently unable to create a coherent, smooth story when given a long narrative text, particularly if the input is not in English.


Although text-to-video models (e.g.,~\cite{singer2022make}) can be used to turn written text into videos, they have some limitations that may make it necessary to use a series of images to tell a story. These limitations include the fact that text-to-video tools are not available in all languages, can be time-consuming, and they may not be able to effectively visualize processes or phenomena described in printed materials such as textbooks, PDFs, reading books, or instruction sheets. 
%

 Therefore, it is important to have alternative visualization tools that can effectively and immediately convey stories or processes without relying on video. These tools should be easy to use and potentially presented in the form of a series of images that tells a story. The use of non-video visualization tools can be particularly useful in situations where video is not a practical or cost-effective solution.


We thus introduce a simple, fast, and cost-effective augmentation to OpenAI's DALL-E, which is capable of visualizing multi-lingual narrative texts, songs, and stories as a coherent sequence of images. This model can process input in the form of a URL (currently available for certain websites) or a text (\texttt{.txt}) file, and also allows for the visualization of stories with specific settings, such as a particular location or artistic style. Although we mainly provide examples using Hebrew text, which presents the additional challenge of being written from right to left (in contrast to English), this model can work with any language that has an ISO-639 identifier. This tool has the potential to be used for educational purposes, such as visualizing children's books, and short stories in school textbooks, and for helping children understand complex songs or stories. 

\section{The Proposed Multi-Lingual Storytelling DALL-E Model} %

\begin{figure*}[t]
\begin{center}
\includegraphics[scale=0.53]{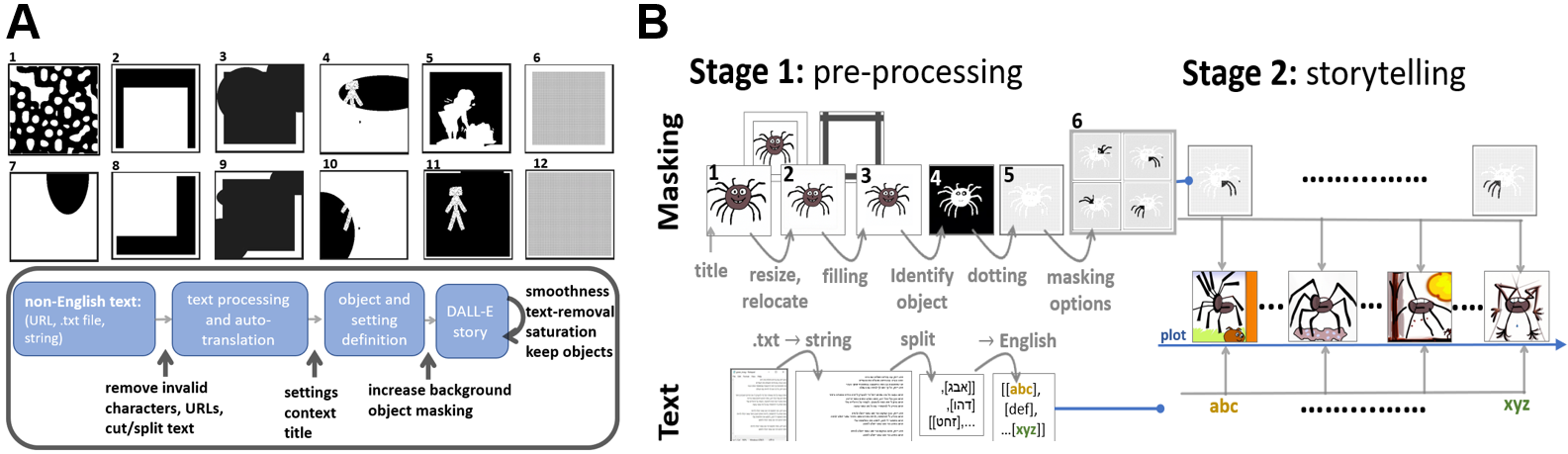}
\end{center}
   \caption{\textit{\textbf{Model Scheme}. \textbf{A:} \textit{Top:} Masking examples. 1 \& 7: Global background change vs local change; 2 \& 8: Edges and Corners; 3 \& 9: Center with applied shapes; 4 \& 10: Object (a man) and a random ellipse; 5 \& 11: Object masking (5: a lady, 11: a man) with fixed edges, with (11) and without (5) a random ellipse; 6 \& 12: dots with fixed edges. \textit{Bottom:} Broad resolution of the model steps and inputs. The model processes and translates non-English text, then defines objects to include, sets constraints and settings, and creates a coherent visual representation of the story. \textbf{B:} Example of pre-processing and masking description for the spider song (the Hebrew version of "Itsi Bitsi Spider"). \cite{spider_song}} }
\label{fig:model}
\vspace{-14 pt}
\end{figure*}

The general scheme of the proposed Multi-Lingual Storytelling DALL-E is presented in Figure~\ref{fig:model}. First, as seen in Figure~\ref{fig:model} \textbf{A}, the model can receive different kinds of non-English text inputs: 1) plain text (string); 2) \texttt{.txt} file; OR 3) a URL address, and can work with any language that has an ISO-639 identifier. It returns a series of coherent images that describe the text content while maintaining consistency of the settings and characters over the frames, where the level of coherency and pace of change is controlled via the model inputs. 
The first step in the process involves cleaning the text by removing invalid characters, signs, HTML signs, URLs, etc. The input text is then divided into a list of short sentences using various text-cutting methods, such as splitting by lines, sentences, and overlapping sentences with a moving window, or stacking overlapping pairs of lines. These methods can be controlled by the user through the model inputs.

Next, the text is automatically translated using the Google Translate API, and the first image in the series is generated based on a set of input options that primarily include: 1) specifying `scene' settings, such as a specific location (by providing an initial settings image background), 2) providing context, such as a specific artistic style, and 3) the ability to maintain focus on a specific figure or object that is expected to be present throughout the story. The selection of these options is determined by the intended result of the visualization and the known structure of the story being visualized. For instance, a fast-paced story such as a song may require a different visualization technique compared to a slower-paced story like a fairy tale.

If the visualization does not require a specific object to be in focus, the model creates a sequence of coherent and continuous frames by generating a new frame for each piece of text by making small modifications to the previous frame using the "edit" feature of the original DALL-E model. The mask used to create each frame based on the previous one is created based on the model inputs. The type of mask used can vary depending on the desired focus, such as preserving the overall structure of the previous frame, maintaining the same setting, or constantly updating or changing a specific area (examples of possible masks are shown in \ref{fig:model}). In addition to this coherency feature, each image is processed through a text-removal algorithm (based on Python's \texttt{keras\_ocr} package and OpenCV inpainting), and the saturation level of each frame is adjusted to match the saturation of the first image (by adjusting the saturation components in the HSV representation of each image).

If the story focuses on a character or object that should remain the same throughout the frames, with only small modifications such as the ability to move (as seen in the "The Story of the Green Man" example in Figure~\ref{fig:green_spider} \textbf{B}), the object is first created based on the story/song title, with a white background, by adding the request "on white background" to the initial DALL-E input prompt. An example of the output of such a request is shown in the top left-most image (Fig.~\ref{fig:model} \textbf{B}, Stage 1, \textbf{1}). The use of the song/story title for this process ensures that no prior knowledge about the story/song content is necessary, allowing the framework to be general enough to be used with any text content without the need for manual input of prior knowledge. 
Subsequently, the object is resized and repositioned to accommodate the inclusion of additional elements later in the story creation, so that the main object does not occupy the entire frame (see Fig.~\ref{fig:model} \textbf{B}, Stage 1, \textbf{2}). To ensure that the resizing of the object is smoothly integrated and consistent with the white background, a DALL-E editing step is applied to fill the gap between the constrained white edges of a white background image and the resized image obtained in the previous steps. This step is particularly important when the original image creation includes an object that is partially cut off, such as a zoomed-in view of a spider or only the top half of a person. In these cases, this step is essential for filling in the missing part of the object rather than cutting it in the middle.
Next, the object is identified by recognizing the non-white pixels (see \textbf{3} to \textbf{4} in Fig.~\ref{fig:model} \textbf{B}), and a dotted masking structure is applied to the background (where the object does not exist) in order to maintain smoothness and coherence in the background while still allowing some flexibility for changes. This is achieved by placing dots at regular intervals (typically every $x$ pixels, where $2\leq x\leq 5$) except for constant $y$-wide edges (typically $y$ is between 10 and 40 for 256x256 images) to enable change while maintaining a coherent yet flexible background. The use of a dotted masking structure allows for a balance between smoothness and flexibility in the background.
Then, to allow for small movements, orientations, or mimicked changes in the look-after main object, a portion of the dominant object is also adjustable in each frame of the story generation (Fig.~\ref{fig:model} \textbf{B}, Stage 1, \textbf{6}). The specific area that is altered is randomly determined, and the user has the option to specify the extent of the change and which areas should remain unchanged. For example, if the face of a person should be preserved, the user can specify that the upper part of the object should not be modified. Additionally, the model is able to make small, global changes to the object by incorporating low-frequency dots in the mask within the object's region (the frequency of these dots is a user-specified model input).
If a location setting is specified for the model (e.g., a constrained initial background at the Johns Hopkins campus), the detected object is placed on the provided setting image. If no location setting is given, the object is placed on a background image generated by DALL-E using the first text piece as an input prompt.
As an aside, we initially attempted to use well-known object detection algorithms trained on widely used datasets (e.g., \cite{redmon2016you} trained on ImageNet \cite{deng2009imagenet}), however, we found that using pre-trained models often resulted in the inability to recognize imaginary or less realistic objects, or objects that do not have a pre-defined label (e.g., "The Green Man"). Training an object detection model from scratch to include such uncommon labels is also not practical as it would take significant computational overhead, which conflicts with the speed and reuse of widely available technology in our proposed framework as one of its main strengths. As a result, we opted not to use this time-consuming approach and to use the practice described above instead.

The process of creating the story in the object detection case is similar to that described previously for the non-object case (namely, continuously generating new frames with continuity constraints using masks). However, the main difference in the object case is that the masks take into account the need to maintain object coherency across frames (see Fig.~\ref{fig:model} \textbf{B}, Stage 2).

\section{Experiments}
\subsection{Object with Free Settings}
\begin{figure*}[t]
\begin{center}
\includegraphics[scale=0.45]{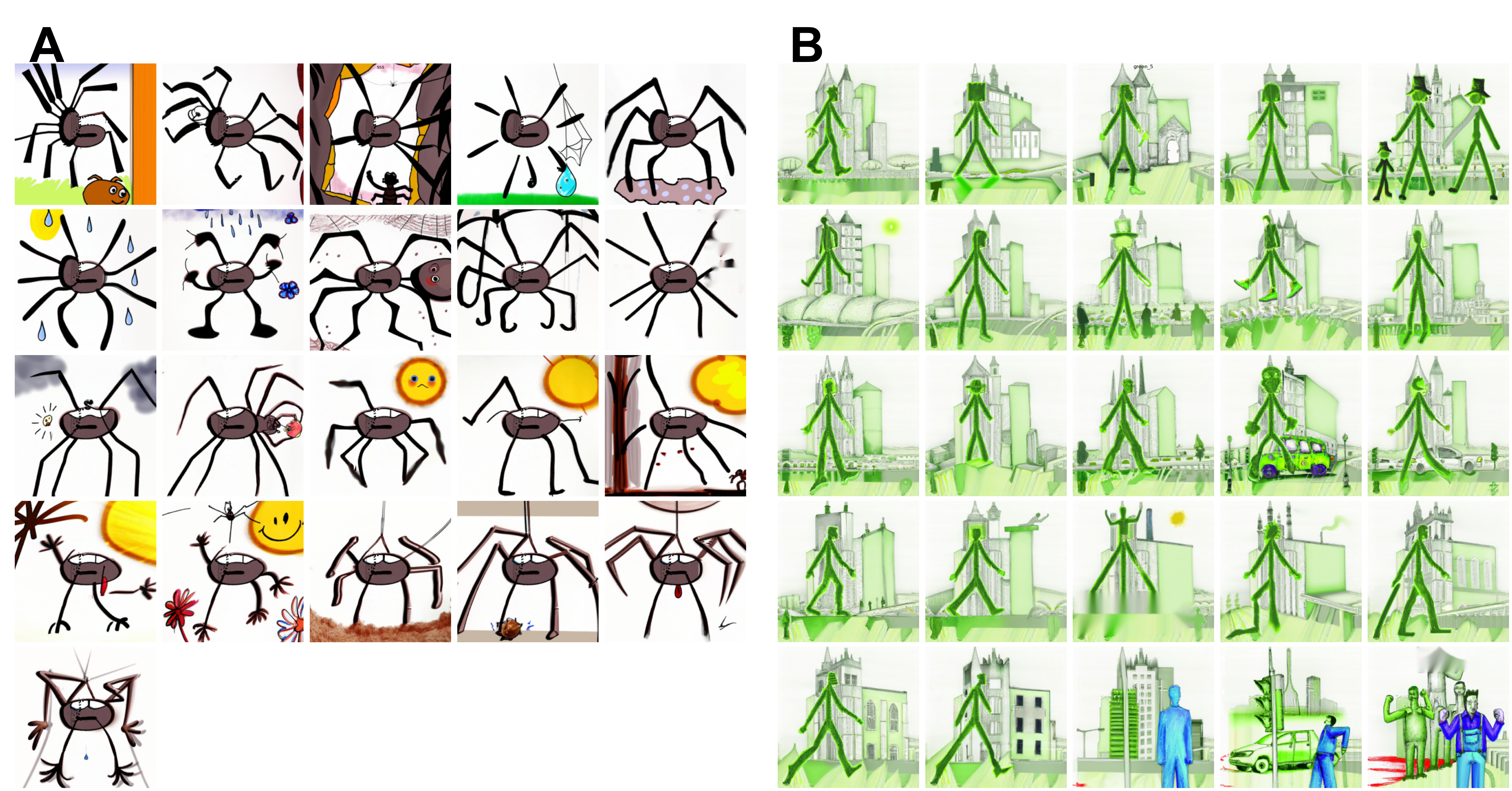}
\end{center}
   \caption{\textit{
   \textbf{Dominant object with free setting examples}.In these examples, the process begins by generating the object (such as a spider or a green man) with the desired background style and extracting and masking it. Subsequent frames are then created with masking constraints that apply to the object and global masking constraints to maintain a consistent background between frames. Between each pair of consecutive frames, a different part of the object is slightly modified in order to maintain coherence within the story (see masking examples in Fig. \ref{fig:model}). \textbf{A:} This visualization presents the Hebrew version of the "Itsi Bitsi Spider" nursery rhyme children's song in a clear and concise manner, with the same spider appearing in each frame.  \textbf{B:} Description of the Hebrew children's song "The Story of the Green Man" \cite{geffen}, that tells about a green man, who lives in a green house, has a green family and drives a green car. At the end of the story, the green man meets a blue man, who happens to be from a different story.}}
\label{fig:green_spider}
\vspace{-14 pt}
\end{figure*}
To create a believable and coherent story, it is important to maintain the same appearance and object consistency over frames. However, this can be a challenge for AI models when the object is moving or interacting with other objects.
In Figure \ref{fig:green_spider}, we use the multi-lingual storytelling DALL-E to show stories with dominant objects that should remain constant across frames, such as animals or humans. Our goal is to maintain consistency in the appearance of a specific object (such as a spider or a green man) across all frames, while still allowing for movement, changes in orientation, and facial expressions. This can be a challenge in general when visualizing stories if we do not pay attention to maintaining the same objects. To demonstrate this, we present the model's results on two songs: 1) the ``Itsy Bitsy Spider'' song using its Hebrew version~\cite{spider_song}, utilizing the model's multi-lingual capabilities, and 2) ``The Story of the Green Man'', a Hebrew children's song about a green man.

In the spider song example, which is illustrated in Figure~\ref{fig:green_spider} \textbf{A}, the model demonstrates its ability to enable the spider to move and change its surroundings while retaining its distinct identity throughout the sequence. The sequence begins with the spider climbing, then raindrops appear, and finally, the sun emerges.

The green man example illustrates the plot of the Hebrew song ``The Story of the Green Man'' (originally called ``Hasipur Al Ha'ish Hayarok''), written by Yehonatan Geffen~\cite{geffen}. The song tells the story of a man who lives in a green city with a green family, drives a green car, and resides in a green house with green windows. In the visual representation shown in Figure ~\ref{fig:green_spider} \textbf{B}, the top row depicts the green man's green house and green family, the second row shows him wearing green shoes and a green hat, the third row shows his green car, and the fourth row shows him moving towards a green flower (column 2) and happily jumping on a beautiful day with the sun shining (column 3). In the fifth row, the green man encounters a blue man on the sidewalk, stops his car, and learns that the blue man is from a different story.

\subsection{Defined Setting}

\begin{figure}[h]
\begin{center}
\includegraphics[width=0.47\textwidth]{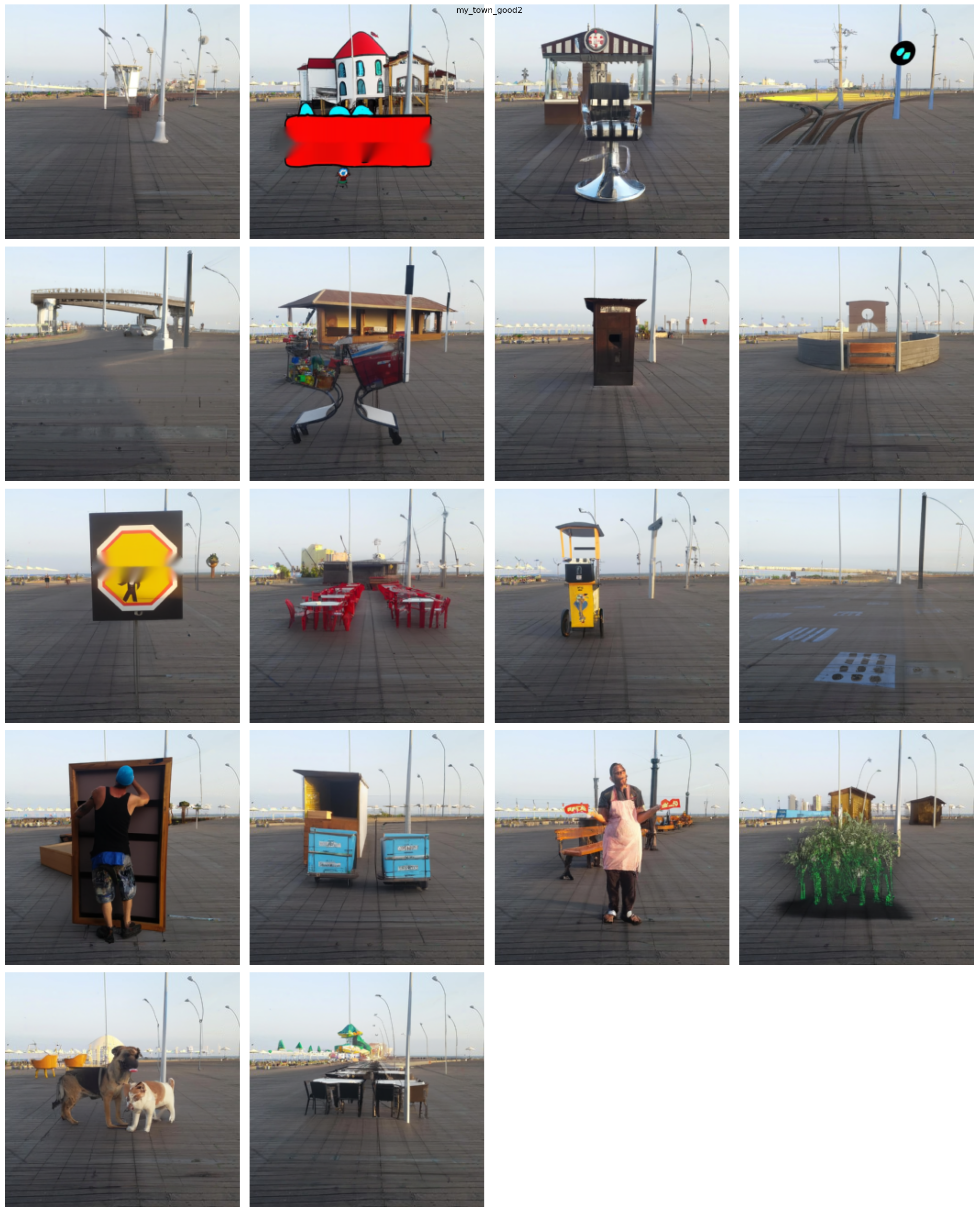}
\end{center}
   \caption{\textit{\textbf{Defined Setting, flexible object}.Description of the Hebrew song ``If You Are Around'', in memorial of Itzhak Klepter, who wrote and composed this song.}}
\label{fig:tlv}
\vspace{-13 pt}
\end{figure}

In Figure~\ref{fig:tlv}, we demonstrate our proposed model's ability to visualize a series of objects within the same setting while maintaining and shaping that setting over the different frames. As an example, we use the model to visualize the song ``If You Are Around''~\cite{klepter}, written and composed by the late Israeli musician Itzhak Klepter, in honor of his memory. The song describes the people and elements in a city, and we apply our proposed DALL-E storytelling model to it with a Tel-Aviv-Port setting, by constraining the visualization frames to a photo taken at the Tel Aviv port.

In the top row, the second column shows houses with red roofs, followed by a hair salon and train tracks in the third and fourth columns (consistent with the song's content). The second row depicts a bridge in the first column, a grocery store in the second column (with a shopping cart in the foreground), and the third column shows a ``mailbox without a window''. In the third row, a stop sign is visible, as the narrator of the song instructs the listener to stop if they see him. Unfortunately, the blurring on the stop sign is an undesired artifact caused by the text removal feature of the model. The fourth column of the third row shows the holes drilled by the smith, and the fourth row shows the smith building a window (first column), along with Mrs. Schmidt (third column), and weeds (fourth column). Dogs and cats appear in the bottom row, all of which are described in the song. 

\subsection{Defined settings and object}
\begin{figure}[h]
\begin{center}
\includegraphics[width=0.47\textwidth]{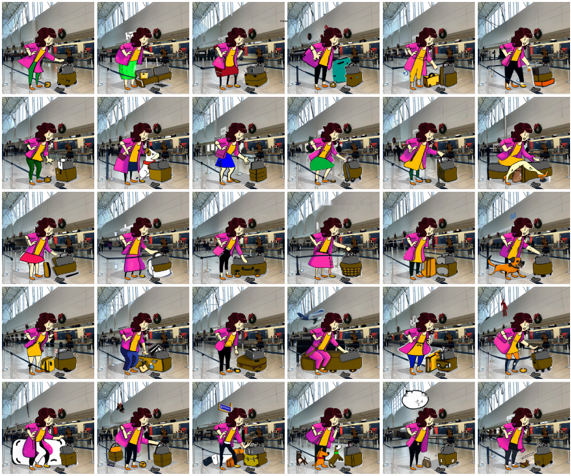}
\end{center}
   \caption{\textit{\textbf{Defined settings and object}. Part of the "Baggage" song (used in its original Russian version) written by Samuil Marshak \cite{samu}.} }
\label{fig:russia}
\vspace{-14 pt}
\end{figure}

In previous experiments, we demonstrated the model's ability to process objects of interest with a flexible initial setting (Fig.~\ref{fig:green_spider}), as well as a constant setting with changing objects (Fig.~\ref{fig:tlv}). However, these examples were all in Hebrew. To further evaluate the model's language adaptability, we present a case study using a part of a Russian children's song, called ``Baggage'' (informal English translation), written by Samuil Marshak~\cite{samu}. For the visualization of this song, in \ref{fig:russia}, \textbf{both} the setting (BWI airport) and an object (a woman checking-in luggage) are defined and kept throughout the story. This example also illustrates the model's ability to handle additional languages and maintain consistency within a given setting and with a dominant human object.



In Figure~\ref{fig:russia}, the top row depicts the woman checking in various objects, including a sofa, a suitcase, baskets, a cardboard box, and a dog (shown in columns 2, 3, 6 of row 1 and columns 1, 2 of row 2). The subsequent frames in rows 2 and 3 show the sofa (row 2, column 6), suitcase (row 3, column 3), and dog again (row 3, column 6), in accordance with the lyrics of the song. These objects continue to be depicted throughout the rest of the visualization in a manner that accurately reflects the descriptions provided in the song.

\subsection{Bibilical Stories with Context}
Biblical stories can be particularly difficult to understand due to the presence of abstract or unrealistic scenarios \cite{feinberg2013teaching, stack2014beyond}. A tool that allows for the visualization of these stories can potentially facilitate understanding and promote learning, particularly for those encountering these stories for the first time.


There have been numerous artistic works throughout history to visualize biblical stories. In the realm of sculpture, Michelangelo's famous ``David'' statue is based on the story of David and Goliath from the Bible. Similarly, the Sistine Chapel ceiling, also created by Michelangelo, contains numerous biblical scenes, including the creation of Adam and the story of Noah's Ark. In cinema, there have also been numerous films that have depicted biblical stories, such as ``Noah'', which is based on the story of Noah's Ark from the Book of Genesis. 

However, manually creating visualizations of biblical stories can be time-consuming and costly. In contrast, automatic AI-based methods offer a promising alternative for generating visualizations quickly and cost-effectively, without the need for expensive resources or prior knowledge of the story. 

To the best of our knowledge, there is no fast, affordable, and user-friendly tool currently available that allows users to create smooth and coherent visualizations of biblical stories using AI, specifically using the original Hebrew versions of the stories.
Using our proposed multi-lingual storytelling DALL-E for visualizing biblical stories not only helps improve understanding of the stories, but also allows for the incorporation of context and familiar locations into the visualization, making the stories more accessible and familiar to children.



\begin{figure}[!h]
\begin{center}
   \includegraphics[width=0.47\textwidth]{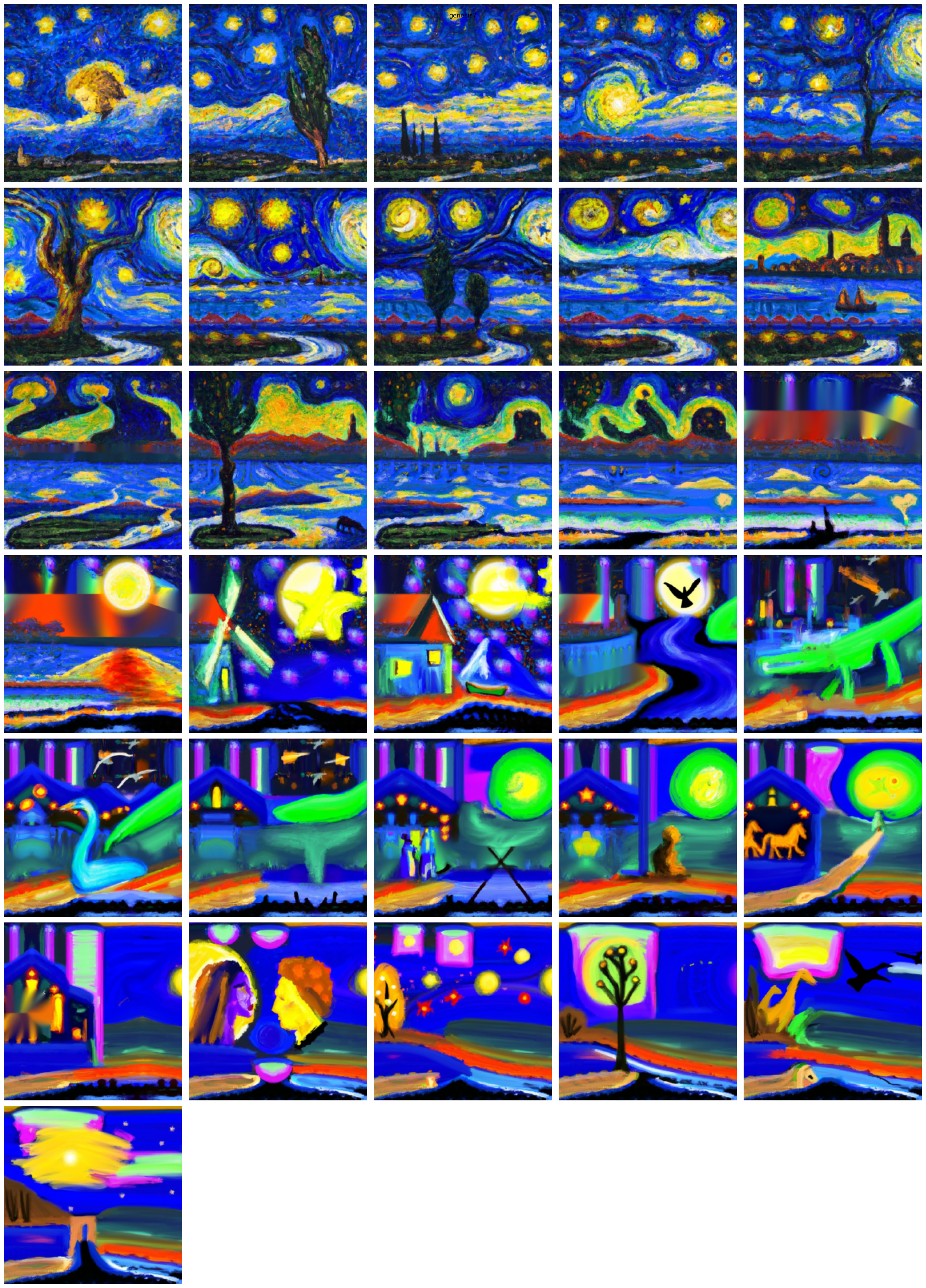}
   \end{center}
   \caption{\textit{\textbf{Biblical Story, under context}.The Book of Genesis Chapter 1 has been visualized using the original Hebrew version and a style constraint of "in van Gogh's The Starry Night style." Familiar motifs from the story are readily identifiable, including the creation of the ``Great Lights'', the animals, and Adam and Eve.}
    }
\label{fig:gens}
\vspace{-14 pt}
\end{figure}

In Figure\ref{fig:gens}, we present the visualization of Genesis Chapter 1 using the multi-lingual storytelling DALL-E framework with the context of ``The Starry Night'' painting by Vincent van Gogh. 
On the first row of Figure~\ref{fig:gens}, one can see a potential visualization of god creating the earth (column 1), along with the creation of light (column 4). In the second row, the separation between the sky and the sea is presented, and the emergence of land is apparent in the second and third rows, along with the creation of herbs and a tree. Then, the creation of the \"two great lights\" as well as the stars is shown in row 4. In rows 4 and 5, the creation of the winged fowls (row 4 column 4) and water animals (row 4 column 5, row 5 columns 1,2). 

An interesting aspect of using the Hebrew version of the text, rather than the formal English version, is that it leads to the visualization of ``great alligators'' rather than ``great whales'' in verse 21. This is because the Hebrew version refers to these creatures as alligators, while the formal English version uses the term ``whales''.
This discrepancy is reflected in the visualization on the rightmost image of the fourth row, which depicts an alligator. 

In the fifth row, fifth column (rightmost image), the creation of land animals can be seen, and in the sixth row, the creation of Adam \& Eve is presented (second column), along with the giving of green herbs and a fruit tree (sixth row, third and fourth columns).

\subsection{Discussion and Future Steps}
In this work, we present a tool for quick and easy visualization of multi-lingual songs and short stories by leveraging the DALL-E text-to-image model to the multi-lingual storytelling case. Our proposed tool is able to handle a variety of scenarios and types of stories, including the incorporation of context and pre-defined settings. We have addressed challenges such as maintaining the same characters and ensuring the smooth and coherent progression of the story. We believe that this tool can be applied to various languages and can visualize difficult-to-understand stories, making it a valuable resource for improving literacy education and story understanding among children. 


For future work, we plan to improve some of the model's features and extend and adjust it to visualize other processes beyond stories, such as scientific phenomena, instructions, or movements. 
In Figure~\ref{fig:russia}, there is an issue with the moderate change in the woman's clothing over time. While the model was able to maintain the woman's identity, it did not sufficiently focus on keeping the same clothing throughout the frames. To address this issue in the future, one potential solution is to adjust the color distribution of relevant objects in the frames to maintain consistency.


While we recognize that the ideal approach to creating a multi-lingual version of DALL-E would be to train the model on every language or to use transfer learning of the existing English model, this would require a significant amount of resources that are not readily available to the average user, and developing such a model would be a lengthy process that would require a large amount of time, money, and computational power. Additionally, collecting data for some less common languages may be challenging. As we strive to eventually attain a multi-lingual DALL-E model that has been trained on every language and performs at the same level as the English version, our current model provides a promising solution for promoting education and equal opportunities for non-English speaking populations. 


With the recent release of the DALL-E API, our approach has been designed to be an accessible, inexpensive, and user-friendly tool for the immediate visualization of stories.  
In the future, we plan to expand this approach to developing an app for kids that allows for quick immediate coherent visualizations of songs and stories with the press of a button, as well as a Chrome extension for visualizing stories from webpages.
Upon publication, we will make this tool available as a pip-installable Python package and will share the code and additional examples on our GitHub repository at https://github.com/NogaMudrik/multi-lingual-storytelling-dall-e.

\subsection{Acknowledgements}
We thank Aya Mudrik, Eva Yezerets, Iuliia Dmitrieva, and Michael Xie, for their helpful feedback and ideas for relevant songs. We are grateful to Google for research credits and to OpenAI for access to the DALL-E API.


\bibliographystyle{plain}
\bibliography{ref}
\newpage 

\end{document}